\documentclass[runningheads]{llncs}
\usepackage[T1]{fontenc}
\usepackage{graphicx}
\usepackage{xcolor}
\usepackage{amsmath}
\usepackage{algorithm}
\usepackage{algorithmic}
\usepackage{booktabs}

\usepackage{amssymb}
\usepackage[hyphens]{url}
\begin{document}
%
\title{Hierarchical Contrastive Learning for Multi-Domain Protein-Ligand Binding}
%
%
\author{Shuo Zhang \and
Rongqi Hong \and
Huifeng Zhang\and
Jian K. Liu}
\authorrunning{S. Zhang et al.}
%
\institute{
University of Birmingham, 
UK 
}
\maketitle              
\begin{abstract}
Predicting protein-ligand binding affinity remains intractable for multi-domain proteins, where inter-domain dynamics govern molecular recognition. Existing geometric deep learning methods typically treat proteins as monolithic static graphs, suffering from rigid-body assumptions and aleatoric noise in flexible regions. To address this, we introduced HCLBind, a self-supervised framework that decouples geometric representation learning from affinity regression. HCLBind leverages a general-to-specific pre-training paradigm on the Q-BioLiP database to learn a robust physical grammar of binding. We propose a novel hierarchical decoy strategy: the model learns local physicochemical constraints through protein coordinate perturbation in single-domain proteins and global conformational geometry through inter-domain rotation in multi-domain complexes. Our hybrid architecture integrates a domain-gated graph attention network and cross-modal attention to explicitly prioritize domain interfaces. Furthermore, we employ LoRA on protein and ligand foundation models, ensuring efficient optimization while preserving evolutionary knowledge. Experiments on PDBBind demonstrate that HCLBind effectively learns discriminative interface features and provides robust uncertainty estimation, overcoming the limitations of standard supervised learning. The code is available at \url{https://github.com/jiankliu/HCLBind}.

\keywords{Multi-Domain Protein  \and Protein-Ligand Binding Affinity Prediction \and Multimodal Learning \and Drug Discovery}
\end{abstract}

\section{Introduction}
Predicting protein-ligand binding affinity is a cornerstone of rational drug discovery, enabling the rapid screening of vast chemical libraries against therapeutic targets. Recently, Graph Neural Networks (GNNs) and geometric deep learning have modeled atomic interactions as topological graphs, explicitly capturing 3D spatial dependencies~\cite{Nguyen2020graphdta,Hua2025mmdgdti}. While effective for single-domain pockets or rigid structures, these methods often operate under a "rigid-body" assumption that oversimplifies the dynamic complexity of biological systems~\cite{Marsh2015structure}.

Multi-domain proteins, which comprise a significant proportion of the proteome, rely on hierarchical inter-domain dynamics rather than just local contacts to govern function~\cite{Gianni2025thefolding}. These dynamics modulate active site accessibility and ligand compatibility, challenging static structural assumptions~\cite{Hansen2025multi}. However, standard multimodal fusion strategies struggle to distinguish between physically valid binding interfaces and geometrically plausible but functionally irrelevant decoys~\cite{Sun2025computer,Kaneriya2025structurenet}. Furthermore, the intrinsic disorder of linker regions introduces aleatoric uncertainty that deterministic models fail to quantify, leading to overconfident predictions on physically implausible or structurally perturbed conformations~\cite{Li2018disordered}.

To address these limitations, we propose HCLBind, a framework tailored for multi-domain complexes. Our approach introduces a hierarchical pre-training strategy that operates at two distinct structural levels:
\begin{enumerate}
    \item Local Level (Physicochemical): We employ a Ligand-Protein Matching (LPM) objective and local noise injection to align atom-level chemical rules and local pocket geometries.
    \item Global Level (Quaternary): We introduce an Interface Decoy Discrimination (IDD) objective using inter-domain rotational perturbations to capture global conformational validity.
\end{enumerate}
To efficiently integrate this hierarchy, we apply Low-Rank Adaptation (LoRA)~\cite{hu2022lora}, following recent parameter-efficient PLM finetuning method~\cite{Zhang2025seqproft}, to pre-trained foundation models (ESMC~\cite{esmteam2024esmc} and MolFormer~\cite{ross2022molformer}). Finally, to address the aleatoric noise inherently introduced by flexible linker regions, we integrate Evidential Deep Learning (EDL)~\cite{amini2020deep} into the downstream regression task.

Our main contributions are:
\begin{itemize}
    \item We formulate protein-ligand affinity prediction for multi-domain proteins as a hierarchical geometric representation problem, rather than treating proteins as static monolithic graphs.
    \item We introduce a hierarchical contrastive pre-training strategy combining local ligand-protein matching with global interface decoy discrimination, enabling the model to learn both physicochemical compatibility and conformational validity.
    \item We integrate domain-aware structural attention, parameter-efficient LoRA adaptation, and evidential regression to improve affinity prediction and uncertainty estimation for flexible or geometrically ambiguous binding cases.
\end{itemize}

\section{Related Work}
Existing deep learning models for protein-ligand affinity, such as GraphDTA~\cite{Nguyen2020graphdta}, primarily rely on 1D sequence or 2D graphs, whereas recent advancements like CASTER-DTA~\cite{Kumar2025casterdta} and DrugBAN~\cite{Bai2023drugban} incorporate 3D equivariant networks and bilinear attention to better capture local structural interactions. Concurrently, multitask frameworks like DeepDTAGen~\cite{Shah2025deepdtagen} jointly optimize affinity prediction and target-aware drug generation to enrich representation learning. To account for spatial complexities, structure-based methods such as TankBind~\cite{lu2022tankbind} partition proteins into independent functional blocks, and generative models like DynamicBind~\cite{Lu2024dynamicbind} capture ligand-induced conformational flexibility via diffusion. Despite these strides, most affinity predictors still treat proteins as monolithic rigid graphs, often neglecting the hierarchical inter-domain dynamics that govern multi-domain targets~\cite{Gohlke2003converging}. Recent domain-aware affinity modeling has begun to address this limitation by explicitly encoding domain semantics and inter-domain interfaces~\cite{zhang2026dagml}. To bridge sequence semantics and structural geometry, multimodal approaches like CL-GNN~\cite{Zhang2025clgnn} utilize contrastive learning~\cite{Singh2023contrastive}, while ProCeSa~\cite{Zhou2025procesa} demonstrates that contrast-enhanced sequence-structure representation learning can improve PLM-based protein property prediction. However, these methods typically optimize global matching objectives without fine-grained structural supervision for domain hierarchies. Furthermore, while uncertainty quantification is critical for molecular discovery, traditional methods like Deep Ensembles~\cite{Lakshminarayanan2017deepensemble} are computationally expensive. Recently, Evidential Deep Learning (EDL)~\cite{amini2020deep,Zhao2025evidti,Soleimany2021evidential} has emerged as a highly efficient deterministic alternative. HCLBind extends these paradigms by integrating domain-aware contrastive learning with EDL to specifically address the structural complexities and aleatoric noise inherent to multi-domain proteins.

\section{Method}

\begin{figure}
\centering
\includegraphics[width=1.0\linewidth]{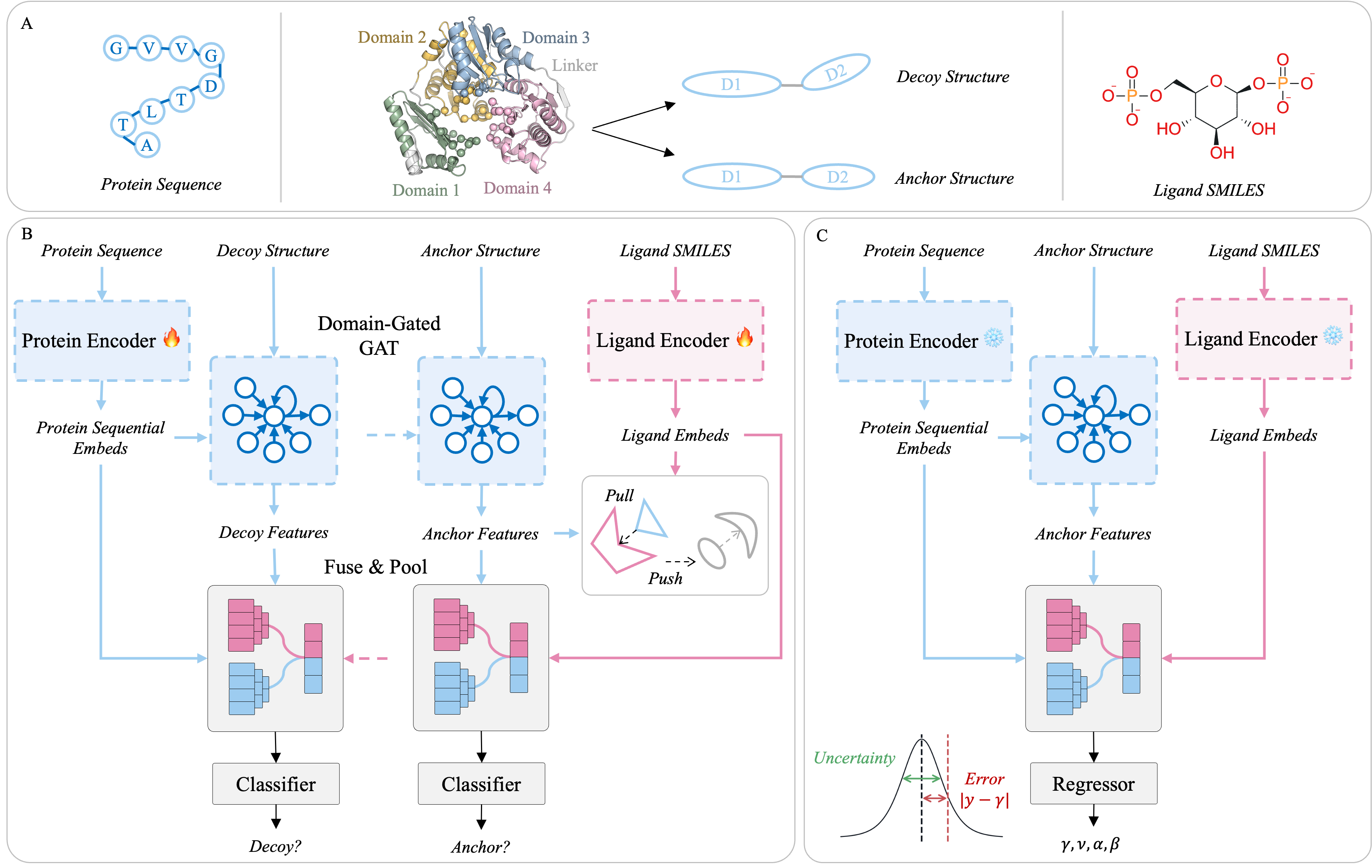}
\caption{The HCLBind Framework. (A) Input Data (B) Pre-training Phase (C) Fine-tuning Phase}
\label{fig:model_arch}
\end{figure}
We propose HCLBind (Hierarchical Contrastive Learning for Binding), a unified framework designed to capture the complex, adaptive geometry of multi-domain protein-ligand interactions. As illustrated in Figure~\ref{fig:model_arch}, HCLBind operates in two phases: a self-supervised pre-training phase that learns to distinguish native domain interfaces from geometric decoys and matches compatible ligands to proteins and a supervised fine-tuning phase that predicts binding affinity with uncertainty quantification.

\subsection{Overview and Problem Formulation}
Let a protein-ligand complex be represented as a tuple $\mathcal{C}=(\mathcal{P},\mathcal{L})$. The protein $\mathcal{P}$ is defined by its amino acid sequences $S=\{s_1,...,s_{N_p}\}$ and its 3D coordinates $X=\{x_1,...,x_{N_p}\}\in\mathbb{R}^{N_p\times3}$. We assume that the protein may consist of multiple conserved domains, denoted by a domain assignment vector $D\in\{0,1,...,K\}^{N_p}$, where $D_i\in\{1,...,K\}$ indicates the domain membership of residue $i$ out of $K$ total domains. The ligand $\mathcal{L}$ is represented by its SMILES string $S_L$.

The primary objective is to learn a joint representation $Z_{complex}$ that captures the physicochemical compatibility and geometric fitness of $\mathcal{L}$ within $\mathcal{P}$.

\subsection{Hierarchical Multi-Modal Encoding}
To effectively model the interplay between sequence semantics and structural geometry, we employ a dual-branch architecture with Low-Rank Adaptation (LoRA) applied to the language model backbones to preserve pre-trained biochemical knowledge while adapting to binding-specific tasks.

\textbf{Domain-Aware Protein Encoding} For sequence representation, we utilize the pre-trained protein language model ESMC to extract residue-level semantic embeddings $H_{seq}\in\mathbb{R}^{N_p\times d_{prot}}$. To capture long-range dependencies and domain-specific geometries in the structural representation, we introduce a Domain-Gated Graph Attention Network. We construct a spatial graph $\mathcal{G}=\mathcal{V},\mathcal{E}$ where nodes are residues and edges are defined by a spatial distance threshold $\tau$. The structural embeddings $H_{struct}$ are computed via an attention mechanism explicitly biased by domain boundaries. For a residue pair $(i,j)$, the attention coefficient $\alpha _{ij}$ is computed as:
\begin{equation}
    \alpha _{ij}=\text{softmax}_{j\in \mathcal{N}(i)}\left(\frac{(W_Qh_i)^\intercal(W_Kh_j)}{\sqrt{d_k}}+\beta _{ij}\right)
\end{equation}
where $h_i$ is the projected hidden state of node $i$. The term $\beta _{ij}$ is a learnable structural bias defined to regulate information flow across domain interfaces. Let $\mathcal{E}$ denote the set of spatial edges in the residue graph, and let $\mathcal{N}(i)=\{j \mid (i,j)\in\mathcal{E}\}$ denote the spatial neighbors of residue $i$:
\begin{equation}
    \beta _{ij}=\begin{cases} 
    b_{interface} &\text{if}\ i,j\in\mathcal{E},\ D_i\ne D_j,\text{and}\ N_{dom}\geq2 \\
    0 & \text{if}\ i,j\in\mathcal{E} \text{ and } D_i=D_j \\
    -\infty &\text{otherwise} 
    \end{cases}
\end{equation}
Here, $b_{interface}$ is a shared and learnable scalar bias applied to inter-domain edges ($D_i\ne D_j$). Given the scarcity of high-quality multi-domain binding data, utilizing a single global scalar rather than a highly parametrized routing network acts as a structural regularizer. It allows the model to uniformly prioritize or dampen cross-domain information flow without overfitting to the specific topologies of the training set. Intra-domain edges ($D_i=D_j$) receive no additional bias (0), while non-neighbor residue pairs are masked by assigning $\beta_{ij}=-\infty$.

\textbf{Ligand Encoding}
We utilize the pre-trained MolFormer model to extract atom-level embeddings $H_{lig}\in\mathbb{R}^{N_l\times d_{lig}}$ from the SMILES string $S_L$. These embeddings are projected via a linear layer to a hidden dimension $d_{hidden}$ matching the protein structural embeddings.

\subsection{Cross-Modal Fusion}
To synthesize the protein and ligand features, we employ a Multi-Head Cross-Attention (MHCA) module. To explicitly model the ligand "scanning" the protein surface, the ligand embeddings $H_{lig}$ serve as queries ($Q$), while the protein structural embeddings $H_{struct}$ serve as both keys ($K$) and values ($V$).
\begin{equation}
\begin{split}
    Z_{complex}=\text{Pool}(\text{MHCA}(Q=H_{lig},K=H_{struct},V=H_{struct}))
\end{split}    
\end{equation}

The final complex representation $Z_{complex}\in\mathbb{R}^{d_{hidden}}$ is obtained via global pooling over the ligand tokens, masked by valid atoms.

\textbf{Self-Supervised Pre-training Strategy}
\label{sec:pretraining-strategy}
We address the scarcity of labeled affinity data via a hierarchical pre-training phase. We employ two complementary objectives designed to teach the model to recognize valid protein-ligand interfaces and compatible molecular pairings.

\textbf{Interface Decoy Discrimination}
This task forces the model to learn valid 3D binding geometries. We generate structural decoys $\tilde{X}$ by perturbing the native coordinates ${X}$ based on the protein's domain architecture: (1) For multi-domain proteins ($N_{dom}\geq 2$), we simulate non-functional conformational states by applying random rigid-body rotations ($15^{\circ}$ and $30^{\circ}$) to a single domain relative to the complex. (2) For single-domain proteins ($N_{dom}=1$), we apply Gaussian noise ($\sigma=1.5$\r{A}) to the protein coordinates to simulate a distorted local geometry causing steric clashes.

The model is trained to distinguish the native complex embedding $Z_{native}$ from the decoy complex embedding $Z_{decoy}$ using a binary classification loss:
\begin{equation}
\begin{split}
    \mathcal{L}_{IDD}=-\frac{1}{2}[\text{log}\sigma(f_{disc}(Z_{native}))+\text{log}(1-\sigma(f_{disc}(Z_{decoy})))]
\end{split}
\end{equation}

where $f_{disc}$ is a projection head mapping the complex embedding to a validity score.

\textbf{Ligand-Protein Matching}
To learn physicochemical compatibility, we employ a contrastive learning objective. Within a mini-batch of $B$ complexes, the matched pair $(\mathcal{P}_i,\mathcal{L}_i)$ is treated as positive, while all other pairs $(\mathcal{P}_i,\mathcal{L}_{j\neq i})$ serve as negatives. We optimize the InfoNCE loss:
\begin{equation}
    \mathcal{L}_{LPM}=-\frac{1}{B}\sum_{i=1}^B\text{log}\frac{\text{exp}(\text{sim}(z_i,z_i')/\tau)}{\sum_{j=1}^B\text{exp}(\text{sim}(z_i,z'_j)/\tau)}
\end{equation}
where $z_i$ and $z_j'$ are normalized projections of the complex embeddings for the $i$-th protein with the $i$-th and $j$-th ligands, respectively. 

The total pre-training objective is
\begin{equation}
    \mathcal{L}_{pre}=\lambda_1\mathcal{L}_{IDD}+\lambda_2\mathcal{L}_{LPM}
\end{equation}

\subsection{Fine-tuning with Evidential Regression}
For the downstream affinity prediction task, we freeze the language model backbones and replace the pre-training heads with an evidential regression head. To address the inherent noise in experimental affinity data and improve model robustness, we employ Evidential Deep Learning (EDL). Instead of predicting a single scalar affinity $\hat{y}$, the network outputs the parameters $\textbf{m}=(\gamma,\nu,\alpha,\beta)$ of a Normal-Inverse-Gamma (NIG) distribution. This models the higher-order probability distribution of the binding affinity, allowing us to quantify both epistemic (model) and aleatoric (data) uncertainty. The network is trained by minimizing a total evidential loss $\mathcal{L}_{evi}$, composed of the negative log-likelihood of the model evidence and a regularization term that penalizes misleading evidence.

\subsection{Datasets}
To support our hierarchical learning objectives, we curated two datasets enriched with Pfam domain annotations. For self-supervised pre-training, we utilized the Q-BioLiP database~\cite{Wei2024qbiolip}, extracting a non-redundant set of complex structures stratified into single- and multi-domain categories. For the downstream affinity prediction task, we employed the PDBbind database (v2020)~\cite{Liu2017pdbbind} with a strict time-based split (testing on complexes released in or after 2019) to prevent data leakage. We also classified test complexes into Single-Domain, Interface, and Linker Binders to evaluate topological robustness. Detailed dataset curation protocols, context window definitions, and stratification statistics are provided in Supplementary Section S1.

\section{Experiment}
\subsection{Experimental Setup}
Model performance was evaluated using Root Mean Squared Error (RMSE), Pearson Correlation Coefficient (PCC), and Concordance Index (C-Index). HCLBind was implemented in PyTorch and optimized via AdamW.

\subsection{Main Results and Ablation Studies}
To evaluate the predictive performance of HCLBind, we benchmarked it against other state-of-the-art models. As summarized in Table~\ref{tab:main_results}, HCLBind consistently outperforms all baselines, achieving the lowest RMSE (1.309) alongside the highest PCC (0.698) and C-Index (0.744). Notably, the most competitive baseline, CL-GNN, also employs a contrastive pre-training paradigm to learn complex representations. However, HCLBind surpasses CL-GNN, which suggests that our hierarchical pre-training approach extracts more discriminative multi-domain interaction features than global contrastive alignment alone.

To further assess the contribution of our proposed components, we conducted an ablation study by systematically excluding core modules (LoRA, LPM, IDD, and EDL) from the full HCLBind framework. The detailed results are also summarized in the bottom half of Table~\ref{tab:main_results}.

\begin{table}
    \centering
    \caption{Main results on PDBbind test set.}
    \label{tab:main_results}
        \begin{tabular}{l|ccc}
        \toprule
        Model Variant                           & RMSE ($\downarrow$)   & PCC ($\uparrow$) & C-Index ($\uparrow$) \\
        \midrule
        DrugBAN~\cite{Bai2023drugban}           & 1.544 & 0.571 & 0.667 \\
        GraphDTA~\cite{Nguyen2020graphdta}      & 1.511 & 0.592 & 0.694 \\
        TankBind~\cite{lu2022tankbind}          & 1.505 & 0.656 & 0.729 \\
        DynamicBind~\cite{Lu2024dynamicbind}    & 1.502 & 0.613 & 0.716 \\
        Caster-DTA~\cite{Kumar2025casterdta}    & 1.484 & 0.640 & 0.702 \\
        DeepDTAGen~\cite{Shah2025deepdtagen}    & 1.454 & 0.613 & 0.699 \\
        CL-GNN~\cite{Zhang2025clgnn}            & 1.330 & 0.681 & 0.725 \\
        \midrule
        HCLBind             & \textbf{1.309} & \textbf{0.698} & \textbf{0.744} \\
        w/o EDL             & 1.371 & 0.671 & 0.726 \\
        w/o LoRA            & 1.561 & 0.630 & 0.705 \\
        w/o LoRA \& EDL     & 1.614 & 0.616 & 0.694 \\
        w/o LPM             & 1.491 & 0.661 & 0.723 \\
        w/o LPM \& EDL      & 1.515 & 0.612 & 0.711 \\
        w/o IDD             & 1.507 & 0.646 & 0.711 \\
        w/o IDD \& EDL      & 1.516 & 0.590 & 0.684 \\
        \bottomrule
        \end{tabular}%
\end{table}

\textbf{Impact of Hierarchical Pre-training Objectives} The analysis of the individual contributions of the two self-supervised pre-training objectives leads to two findings:
\begin{itemize}
    \item Geometric Validity (IDD): Excluding the IDD objective significantly degrades performance (RMSE increases to 1.507), confirming that explicit decoy discrimination is essential for learning precise structural binding constraints.
    \item Physicochemical Compatibility (LPM): Similarly, excluding the LPM objective leads to an increase in RMSE to 1.491. This indicates that the contrastive alignment between ligand and protein pockets is important in learning the chemical rules of interaction. The full model outperforms both single-objective variants, demonstrating that geometric validity and chemical compatibility are complementary features that should be learned jointly.
\end{itemize}

\textbf{Effectiveness of Evidential Regression}
The model trained with evidential loss consistently outperforms its deterministic counterparts across all ablation variants (see Table~\ref{tab:main_results}). Beyond enabling uncertainty quantification, the evidential loss acts as an effective regularizer during training, preventing the model from becoming overconfident on noisy data points and improving generalizability.

\textbf{Importance of Parameter-Efficient Adaptation}
Finally, we investigated the role of LoRA in fine-tuning the large protein language model backbone. Freezing language model parameters without LoRA sharply increases RMSE to 1.561. This confirms that LoRA is essential for adapting high-level sequence semantics to binding tasks, avoiding both the overfitting and catastrophic forgetting associated with limited affinity data.

\subsection{Uncertainty Quantification and Reliability Analysis}
\begin{figure}
\centering
\includegraphics[width=1.0\linewidth]{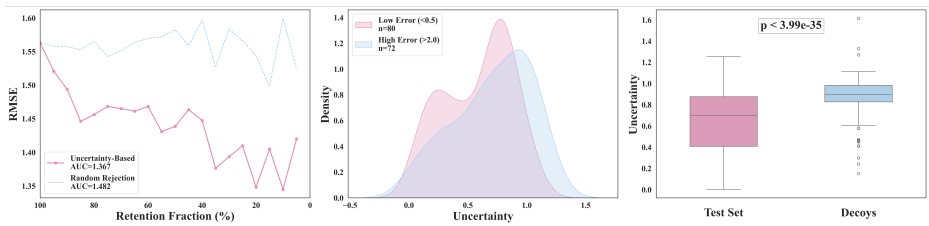}
\caption{Uncertainty quantification and reliability analysis. Left: Error-retention curves. Middle: Density distribution of uncertainty scores. Right: Epistemic uncertainty comparison.}
\label{fig:uncertainty}
\end{figure}
Reliable uncertainty estimation is important for high-stakes drug discovery. HCLBind employs evidential deep learning to quantify epistemic uncertainty, enabling the detection of structurally anomalous inputs and ambiguous binding interfaces. We evaluated the quality of these uncertainty estimates through three distinct analyses, as illustrated in Figure~\ref{fig:uncertainty}.

\textbf{Confidence-Based Error Rejection}
We first assessed whether the estimated uncertainty correlates with prediction error using a sparsification error analysis. We sorted the test set predictions by their confidence (defined as the inverse of the predicted uncertainty) and iteratively removed the most uncertain samples. We then calculated the RMSE of the remaining data at various retention fractions (100\% to 5\%). As shown in the Figure~\ref{fig:uncertainty} left panel, the uncertainty-based rejection curve (solid pink line) exhibits a monotonic decrease in RMSE as uncertain samples are removed. In contrast, the random rejection baseline (dashed blue line) fluctuates without a clear downward trend. The uncertainty-based rejection strategy achieves a lower AUC (1.367) compared to random rejection (1.474), confirming that prediction error correlates with estimated uncertainty. This calibration may support filtering of low-confidence predictions in virtual screening.

Furthermore, density distributions of uncertainty scores (Figure~\ref{fig:uncertainty}, middle panel) reveal that the evidential head skews toward higher uncertainty for high-error samples (>2.0) compared to low-error samples (<0.5), indicating the model uncertainty is positively associated with larger prediction errors.

\textbf{Sensitivity to Geometric Invalidity}
Rather than evaluating chemical out-of-distribution generalization, we sought to determine if the evidential regression head effectively internalized the geometric constraints learned during the IDD pre-training phase. We evaluated HCLBind's epistemic uncertainty distribution on the valid PDBbind test set against structurally disrupted decoys generated via the interface perturbation strategy. If the pre-training was successful, the evidential head should autonomously assign high epistemic uncertainty to these physically implausible geometries.

As shown in the right panel of Figure~\ref{fig:uncertainty}, the model assigns higher uncertainty to the disrupted decoys (median$\approx$0.9) compared to the valid native complexes (median$\approx$0.7). A Mann-Whitney U test confirms this difference is statistically significant ($p<3.99\times10^{-35}$). This result suggests that IDD improves sensitivity to geometrically disrupted inputs. When presented with a decoy that violates these learned geometric constraints, HCLBind correctly flags it as an epistemic unknown rather than making a confident, erroneous affinity prediction.

\subsection{Performance Analysis by Binding Topologies}
\begin{figure}
\centering
\includegraphics[width=1.0\linewidth]{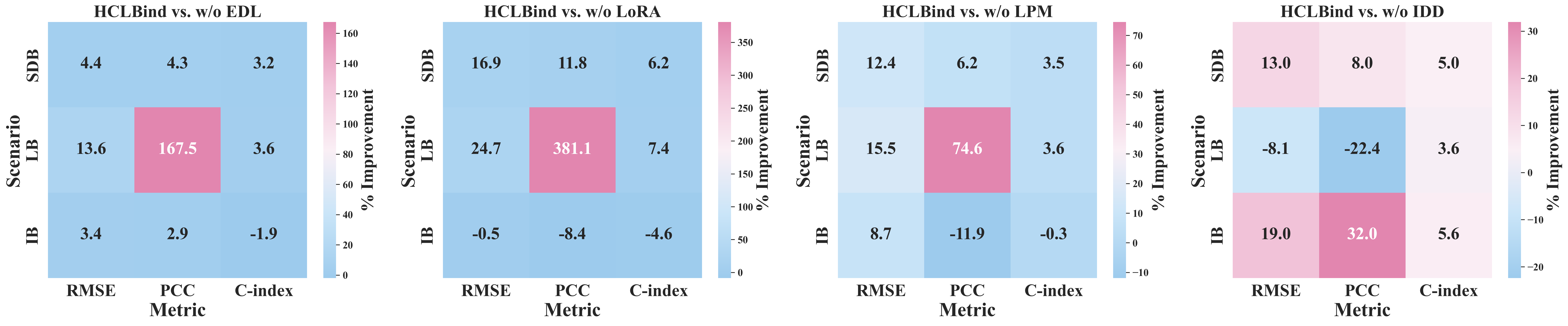}
\caption{Relative performance gain (\%) of HCLBind compared to variants across binding topologies.}
\label{fig:ablation_types}
\end{figure}
As mentioned in Supplementary Section S1.3, we stratified the test set into three categories: Single-Domain Binders (SDB), Interface Binders (IB), and Linker Binders (LB). Figure~\ref{fig:ablation_types} visualizes the relative performance gains of the full HCLBind model compared to its variants.

\textbf{Impact of Evidential Regression.} EDL consistently improves performance across all metrics for SDB and IB, with gains ranging from 2.9\% to 4.4\%. The most dramatic improvement is observed in the LB category, where the PCC improves by 167.5\%. Linker regions are intrinsically flexible and disordered, leading to noisy structural data. Standard regression models likely overfit to this noise, whereas EDL, by modeling the aleatoric uncertainty, allows the network to down-weight unreliable signals from these flexible regions, resulting in significantly more robust correlation.

\textbf{Impact of LoRA.} Introducing LoRA substantially improves performance on SDB and LB categories (LB PCC +381.1\%), confirming its ability to adapt language models to physicochemical binding context. However, the slight performance drop on IB suggests that sequence-based LoRA adaptation may prioritize intra-domain features over the quaternary interactions required for interface binding.

\textbf{Impact of Ligand-Protein-Matching.} The LPM pre-training objective provides consistent gains across SDB and LB categories. Notably, in the IB category, while RMSE improves by 8.7\%, the PCC drops (-11.9\%). While LPM helps the model learn the general fit (lowering absolute error), it may struggle to rank the affinities of complex interface binders where geometric complementarity (handled by IDD) is more important than pure chemical compatibility.

\textbf{Impact of Interface Decoy Discrimination.} For IB, removing IDD causes a severe drop in performance. This validates that explicitly distinguishing native geometries from decoys enables HCLBind to recognize the precise structural clefts necessary for interface binding. Conversely, removing IDD improves performance for LB, likely because rigid geometric constraints do not apply to disordered regions. This highlights that HCLBind's hierarchical pre-training specializes the model for structured domains while the EDL component handles the uncertainty of flexible regions.

\section{Conclusion}
In this work, we introduced HCLBind to address the limitations of rigid-body assumptions in predicting protein-ligand binding affinity, particularly for multi-domain proteins. By implementing a hierarchical pre-training paradigm, we successfully decouple the learning of geometric structural constraints from physicochemical compatibility. Our ablation studies confirm that the Interface Decoy Discrimination objective is critical for recognizing valid inter-domain geometries, while the Ligand-Protein Matching objective ensures accurate chemical alignment.

Furthermore, Evidential Deep Learning successfully mitigated noise from disordered linkers. This allows HCLBind to identify and down-weight unreliable predictions on geometrically invalid structural decoys, a capability lacking in standard deterministic regressors. Finally, LoRA effectively adapts foundation models to the binding task without catastrophic forgetting of evolutionary knowledge. HCLBind offers a reliable framework for rational drug discovery against challenging multi-domain targets.

While this study establishes the fundamental utility of hierarchical contrastive pre-training and evidential regression for multi-domain proteins, we acknowledge certain limitations. Our current empirical evaluation primarily utilizes CASTER-DTA as a representative equivariant baseline and focuses heavily on internal model ablations to isolate the effects of our proposed modules. Future work will expand our comparative analysis to a broader suite of state-of-the-art structure-based and sequence-based baselines, and evaluate the model's chemical generalization on independent, novel-scaffold datasets.

\bibliographystyle{splncs04}
\bibliography{references}

\end{document}